\documentclass[letterpaper, 10 pt, conference]{ieeeconf}
\IEEEoverridecommandlockouts
\usepackage{graphics}
\usepackage{epsfig}
\usepackage{amsmath}
\usepackage{amssymb}

\usepackage{booktabs}
\usepackage{multirow}
\usepackage{subcaption}
\usepackage{graphicx}

\title{\LARGE \bf
SEG-Parking: Towards Safe, Efficient, and Generalizable Autonomous Parking via End-to-End Offline Reinforcement Learning
}
\author{Zewei Yang, Zengqi Peng, and Jun Ma, \textit{Senior Member, IEEE}
\thanks{Zewei Yang and Zengqi Peng contributed equally to this work. \textit{(Corresponding Author: Jun Ma.)}}
\thanks{Zewei Yang and Zengqi Peng are with the Robotics and Autonomous Systems Thrust, The Hong Kong University of Science and Technology (Guangzhou), Guangzhou 511453, China (e-mail: zyang363@connect.hkust-gz.edu.cn; zpeng940@connect.hkust-gz.edu.cn). }
\thanks{Jun Ma is with the Robotics and Autonomous Systems Thrust, The Hong Kong University of Science and Technology (Guangzhou), Guangzhou 511453, China, and also with the Cheng Kar-Shun Robotics Institute, The Hong Kong University of Science and Technology, Hong Kong SAR, China (e-mail: jun.ma@ust.hk).} 
}

\begin{document}

\maketitle
\thispagestyle{empty}
\pagestyle{empty}

\begin{abstract}
Autonomous parking is a critical component for achieving safe and efficient urban autonomous driving. However, unstructured environments and dynamic interactions pose significant challenges to autonomous parking tasks. To address this problem, we propose SEG-Parking, a novel end-to-end offline reinforcement learning (RL) framework to achieve interaction-aware autonomous parking. Notably, a specialized parking dataset is constructed for parking scenarios, which include those without interference from the opposite vehicle (OV) and complex ones involving interactions with the OV. Based on this dataset, a goal-conditioned state encoder is pretrained to map the fused perception information into the latent space. Then, an offline RL policy is optimized with a conservative regularizer that penalizes out-of-distribution actions. Extensive closed-loop experiments are conducted in the high-fidelity CARLA simulator. Comparative results demonstrate the superior performance of our framework with the highest success rate and robust generalization to out-of-distribution parking scenarios. The related dataset and source code will be made publicly available after the paper is accepted.
\end{abstract}

\section{INTRODUCTION}
Autonomous driving has advanced markedly in recent years, yet highly complex scenarios such as parking remain challenging \cite{zhao2024survey}. Unlike typical on-road driving tasks, parking often occurs in unstructured environments, such as open parking lots or curbside spaces, where clear lane markings could be absent \cite{li2021optimization}. Drivable corridors in these settings are often narrow, severely limiting the maneuvering freedom of the vehicle. Moreover, even within these confined spaces, the ego vehicle (EV) is required to dynamically interact with other moving vehicles or oncoming traffic \cite{leu2022autonomous}. Therefore, it is indispensable to develop an autonomous parking system capable of safe and efficient operation under such constraints.

Existing autonomous parking strategies can be broadly grouped into rule-based, optimization-based, and learning-based approaches. Rule-based methods are usually built on vehicle kinematic models and predefined motion primitives, with typical techniques including geometric curve planners and graph search algorithms \cite{sheng2021autonomous}. While these methods can produce collision-free paths in various standard parking scenarios, their reliance on hand-crafted primitives and heuristics limits their generalizability to diverse environments. Optimization-based methods often treat the coarse path obtained by rule-based methods as an initial solution and refine it by solving a constrained trajectory optimization problem \cite{li2021optimization, guo2025fast}. By explicitly accounting for physical limits and safety constraints, such methods can provide trajectories with guaranteed feasibility and optimality rationales. However, their performance depends on high-precision perception for accurate constraint specification. Moreover, purely numerical optimization can incur notable computation time when informed initial guesses are unavailable \cite{dai2018improving}. Learning-based approaches leverage data-driven models to learn parking policies. A prominent paradigm is imitation learning (IL), which trains a policy to mimic expert demonstrations, aiming to reproduce human-like maneuvers \cite{codevilla2018end}. Nevertheless, the performance of IL is bounded by the quality and coverage of expert data, which is particularly concerning given the scarcity of large-scale parking maneuver datasets. In addition, IL is prone to causal confusion \cite{de2019causal}, meaning that it may learn spurious correlations rather than true causal factors. As a result, when the deployment state distribution deviates from the training distribution, its performance can degrade sharply, particularly in interactive parking scenarios \cite{kumar2022should}. In contrast, reinforcement learning (RL) offers an alternative in which policies are optimized to maximize cumulative return over action sequences \cite{kiran2021deep,peng2025bilevel,peng2024reward}. Recent studies \cite{jiang2025hope} show that RL can synthesize diverse and effective parking maneuvers, thereby achieving competitive performance. However, training RL policies from scratch typically entails extensive trial-and-error interaction \cite{gulcehre2020rl}. This is especially severe for parking scenarios, where narrow drivable corridors increase the frequency of failed attempts and markedly reduce sample efficiency. Furthermore, exploration remains challenging in large action spaces for parking, especially for long-horizon tasks where delayed reward assignment leads to inefficient training \cite{tang2025deep}.

\begin{figure*}[t]
  \centering
  \includegraphics[width=\textwidth]{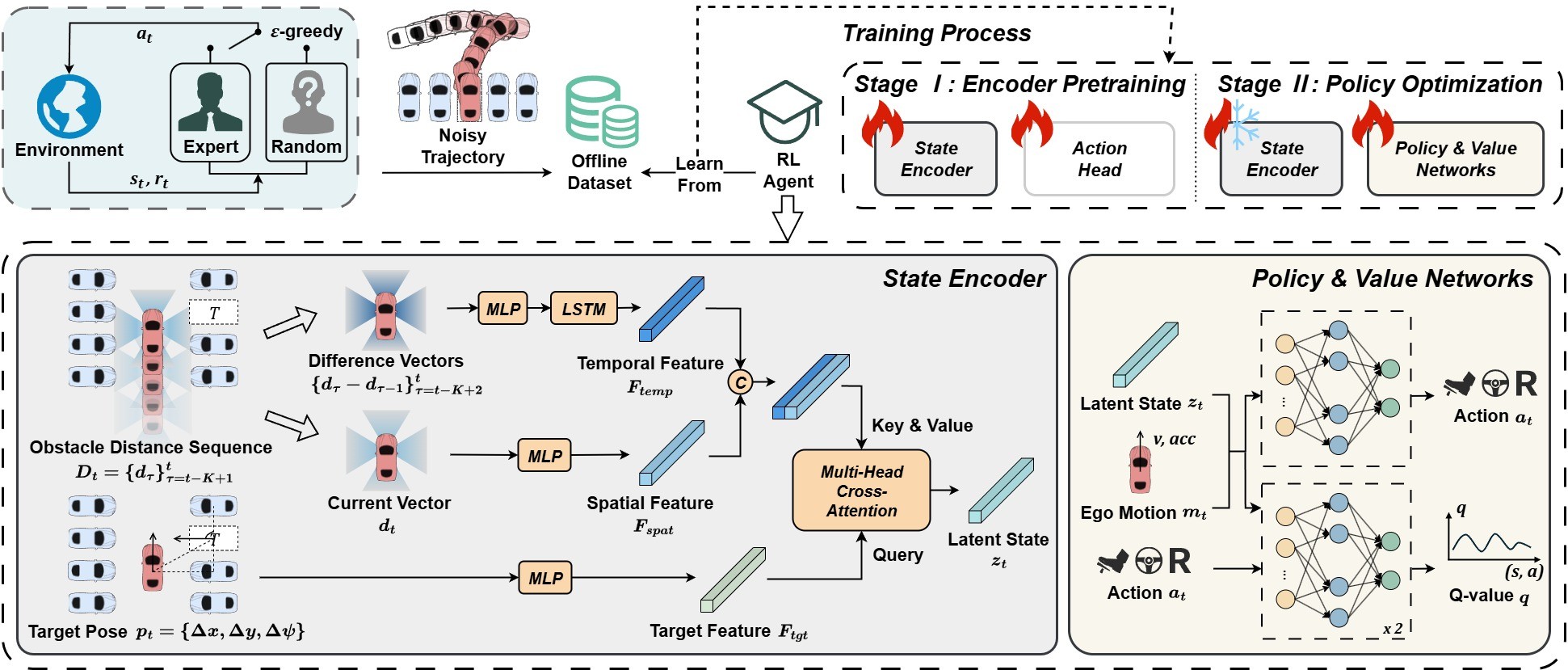}
  \caption{Overview of the proposed offline RL framework for autonomous parking.}
  \label{fig:framework}
\end{figure*}

Motivated by these challenges, we propose an end-to-end offline RL framework for autonomous parking, which has been extensively evaluated in the high-fidelity CARLA simulator. The pipeline of the proposed framework is shown in Fig.~\ref{fig:framework}. The dataset is collected by an expert policy with injected noise in interactive parking scenarios, in which a potential opposite vehicle (OV) interacts with the EV. The overall architecture comprises a state encoder that fuses temporal perception information with the target pose, and a downstream policy that outputs low-level control commands. To facilitate efficient learning, the state encoder is first pretrained to distill underlying behavior patterns within the dataset. Then, the policy is optimized with conservative regularization to mitigate unsafe extrapolation. In summary, our main contributions are:

\begin{itemize}
\item An innovative end-to-end offline RL framework is proposed for interaction-aware driving in unstructured and dynamic parking scenarios. It significantly improves the driving safety, task efficiency, and generalizability.
\item A dataset is constructed for autonomous parking tasks that focus on interaction-aware scenarios. It comprises two key types of scenarios, including those without the OV and those involving complex interactions with the OV. 
\item Extensive experiments are conducted in the CARLA simulator, which demonstrate the superior parking performance and excellent generalization capability of our framework. 
\item The dataset for interaction-aware autonomous parking tasks and the related source code of SEG-Parking will be made publicly available.
\end{itemize}

\section{RELATED WORKS}
\subsection{Rule-based and Optimization-based Parking}
Early autonomous parking systems mainly relies on explicit geometric patterns or systematic search strategies for path planning \cite{sheng2021autonomous}. Geometric planners construct feasible paths using predefined curve templates. For example, Dubins curves \cite{dubins1957curves} yield the shortest path for a forward-driving car with bounded curvature, and Reeds–Shepp curves \cite{reeds1990optimal} extend them by allowing reversing maneuvers. Based on these canonical curve families, various parking planners have been developed with further improvements such as modified curvature transitions \cite{kim2014auto} and multi-segment maneuvers \cite{du2014autonomous} to better fit parking scenarios. Search-based planners have also been widely explored. A prominent example is Hybrid A* \cite{dolgov2010path}, which extends classic A* into the continuous state space of the vehicle via feasible motion primitives. Originally deployed in the DARPA Urban Challenge, Hybrid A* has since been adapted to parking with enhanced heuristics \cite{sedighi2019guided} and multistage search \cite{sheng2021autonomous}. Both geometric and search-based methods leverage rule-based priors to synthesize drivable collision-free paths and have demonstrated effectiveness in most simple scenarios. However, they typically require scenario-specific tuning and could struggle in highly complex environments with irregular obstacles or tight spaces \cite{zips2016optimisation}.

A complementary line of work formulates parking as a constrained optimal control problem (OCP). 
In this formulation, the surrounding obstacles are explicitly encoded as inequality constraints that enforce collision avoidance. For example, a lightweight iterative framework \cite{li2021optimization} is proposed to reconstruct the drivable corridor around irregularly placed obstacles and solve a low-dimensional OCP in each iteration. Similarly, the recent TPCKC \cite{guo2025fast} treats the violated vertex-to-polytope constraints as key constraints and iteratively solves a reduced OCP. While such optimization-based planners can generate trajectories with provable constraint satisfaction, they inherently rely on a well-shaped initial guess to maintain computational efficiency. If the initial guess is unavailable, they can become computationally demanding as scenarios grow more intricate \cite{dai2018improving}.

\subsection{Learning-based Parking}
Learning-based parking approaches utilize neural networks to model complex driving behaviors, bypassing the need for predefined rules and numerical optimization \cite{chen2024end}. Within this paradigm, IL leverages the expert demonstration data to train a policy that directly mimics the behavioral patterns. For instance, an encoder–decoder network coupled with a long short-term memory (LSTM) network, fed by monocular images from front and rear mounted cameras, iteratively estimates steering angle and gear status \cite{rathour2018vision}. Another vision-based method trains a convolutional neural network that maps rear-view images to steering and velocity \cite{li2018end}. 
Recently, transformer-style architectures that fuse multi-view images with target context are used to learn end-to-end parking policies \cite{yang2024e2e, li2024parkinge2e}, thereby streamlining the pipeline for practical deployment. Nevertheless, IL-based pipelines are vulnerable to covariate shift, limiting generalization to states outside the training distribution.

In contrast, RL casts parking as a sequential decision-making problem and optimizes behavior to maximize expected cumulative reward. The Monte Carlo tree search approach \cite{song2022time} has been explored to generate approximate time-optimal path-planning results for parallel parking. Building on this insight, a hierarchical planner \cite{yuan2023hierarchical} broadens its applicability by coupling a high-level deep deterministic policy gradient policy that sketches coarse reference paths with a low-level refinement module. More recently, HOPE \cite{jiang2025hope} integrates an RL policy with Reeds–Shepp curves as a hybrid policy to handle more diverse parking scenarios. However, such methods output waypoints that require a downstream controller, which often introduce tracking deviations in complex scenarios \cite{yang2024e2e}. Moreover, their reliance on extensive online interactions with the environment makes training time-consuming, especially for long-horizon tasks \cite{tang2025deep}. These considerations motivate us to focus on an end-to-end offline RL solution.

\section{PRELIMINARY}
We formulate the autonomous parking task as a Markov Decision Process (MDP), which can be defined formally as $ \mathcal{M} := \{\mathcal{S}, \mathcal{A}, \mathcal{P}, \mathcal{R}, \gamma \} $. Here, $ \mathcal{S} $, $ \mathcal{A} $, $ \mathcal{P} $, $ \mathcal{R} $, and $ \gamma $ denote the state space, action space, state transition dynamics, reward function, and discount factor, respectively.

\textbf{State Space $\mathcal{S}$}: In our MDP, each state $s_t \in \mathcal{S}$ at time step $t$ captures the perception of the EV regarding its surroundings and its current motion relative to the parking target. It comprises three components:

\begin{itemize}
\item \textbf{Obstacle Distance Sequence $D_t$:} 
A stack of $K$ consecutive obstacle distance vectors:
\begin{equation}
D_t = \{ d_{\tau} \}_{\tau=t-K+1}^{t},
\end{equation}
where $d_{\tau}$ is a vectorized representation derived from the LiDAR point cloud. In any $d_{\tau}$, the $i$-th element $d_{\tau}[i]$ denotes the distance to the nearest obstacle along angle $\psi = i \cdot \delta_{\psi}$ in the egocentric coordinate frame, given the angular resolution $\delta_{\psi}$. This sequence provides short-term temporal context.
\item \textbf{Target Pose $p_t$:} The current position and orientation of the EV relative to the target slot, represented as $\Delta x$, $\Delta y$, and $\Delta \psi$ in the egocentric coordinate frame.
\item \textbf{Ego Motion $m_t$:} The current kinematic status of the EV, including its linear velocity $v$ and acceleration $\dot{v}$.
\end{itemize}

\textbf{Action Space $\mathcal{A}$}: Each action $a_t \in \mathcal{A}$ for parking is represented as a 3-element tuple $(a_1,a_2,a_3)$, where $a_1$, $a_2$, and $a_3$ correspond to the acceleration, steering angle, and gear, respectively. For generality, the action range for each dimension $i$ is defined as $\mathcal{A}_i = [-1, 1]$. 

In practice, each dimension is discretized into $N_i$ equally spaced atomic actions, given by $\mathcal{A}_i^{(N_i)} = \left\{ \frac{2j}{N_i-1} - 1 \right\}_{j=0}^{N_i-1}$. Thus, the complete action space is:
\begin{equation}
\mathcal{A} = \{(a_1, a_2, a_3) \mid a_i \in \mathcal{A}_i^{(N_i)}\;\text{for}\;i=1,2,3\},
\end{equation}
where $N_1$ and $N_2$ are determined by the desired resolution of the acceleration and steering angle, and $N_3 = 2$ indicates that the vehicle can only select forward or reverse gear.

\textbf{State Transition Dynamics $\mathcal{P}(s_{t+1}|s_t, a_t)$}: The transition function defines the state transitions given the action of the agent, adhering to the Markov property. It is determined by the environmental dynamics and is unknown to the RL agent.

\textbf{Reward Function $\mathcal{R}$}: Each reward $r_t \in \mathcal{R}$ is designed to encourage progress towards the target slot and successful completion of the parking maneuver, while penalizing collisions and unsafe behaviors. Details will be introduced in Section~\ref{sec:policy optimization}.

\textbf{Discount Factor $\gamma$}: Future rewards are accumulated with a discount factor $\gamma \in (0, 1)$.

\section{METHODOLOGY}
\subsection{Dataset Construction}
In offline RL, policies are usually trained from a fixed dataset, typically collected by a predefined behavior policy. To efficiently synthesize reliable parking maneuvers, we adopt a classical planning-and-control stack as the expert policy. The Hybrid A* planner \cite{dolgov2010path} first generates a collision-free path to the target parking slot, which is then tracked by a finite-horizon linear quadratic regulator (LQR) \cite{kalman1960contributions} to produce an action proposal $\widetilde{a}_t$.

Given that mild stochasticity is common in real-world parking maneuvers \cite{kumar2020conservative}, we inject noise into the expert policy following \cite{fujimoto2019benchmarking,gulcehre2021regularized}. Specifically, in state $s_t$, there is a small probability $\epsilon$ of overriding the expert action $a^*$ with a random exploratory action $a'$. The resulting noisy behavior policy $\beta_{\text{noisy}}$ is formally defined as:
\begin{equation}
\resizebox{\linewidth}{!}{$
\beta_{\text{noisy}}(s_t) =
\begin{cases}
a^*, a^* = \operatorname{arg\;min}_{a\in\mathcal{A}} \lVert a - \widetilde{a}_t \rVert_2 & \text{with probability } 1-\epsilon,\\
a', \;a' \sim \mathcal{U}(\mathcal{A}) & \text{with probability } \epsilon,
\end{cases}
$}
\end{equation}
where $\operatorname{arg\;min}_{a\in\mathcal{A}} \lVert a - \widetilde{a}_t \rVert_2$ selects the action in $\mathcal{A}$ that is closest to the proposal $\widetilde{a}_t$ in Euclidean distance; $\mathcal{U}(\mathcal{A})$ denotes the discrete uniform distribution over $\mathcal{A}$; $\epsilon \in [0,1]$ is the exploration probability.
This $\epsilon$-greedy strategy yields trajectories that are predominantly expert-driven but occasionally exploratory.
We expect the RL agent to learn an effective parking policy from such suboptimal demonstrations by concatenating the successful segments.

After the EV executes the selected action $a_t = \beta_{\text{noisy}}(s_t)$, the next state $s_{t+1}$ is obtained from the sensor streams and onboard localization of the EV. All trajectories are stored as sequences of transitions $(s_t, a_t, r_t, s_{t+1}, \mathbb{I}\{ s_{t+1} \in \mathcal{S}_{\text{parked}}\})$, where $\mathbb{I}\{ s_{t+1} \in \mathcal{S}_{\text{parked}}\}$ is an indicator function that equals $1$ if the vehicle has reached a successfully parked configuration and $0$ otherwise.

\subsection{Network Architecture}
\paragraph{State Encoder} 
To generate a compact latent state representation, we design a state encoder $\phi$ that fuses the obstacle distance sequence with the target pose.

The obstacle distance sequence $D_t \in \mathbb{R}^{L \times K}$ is first temporally differentiated to emphasize motion cues, yielding a stack of difference vectors $\Delta D_t \in \mathbb{R}^{L \times (K-1)}$, where $L$ denotes the length of each vector determined by the angular resolution. This stack is processed by a multi-layer perceptron (MLP) followed by an LSTM, producing a temporal feature $F_{temp} \in \mathbb{R}^{L_t}$ that summarizes recent environmental dynamics. In parallel, the current distance vector $d_t$ is embedded by an MLP, producing $F_{spat}\in\mathbb{R}^{L_s}$ for the instantaneous spatial layout. Separately, the target pose $p_t$ is passed through a feedforward projection to obtain $F_{goal} \in \mathbb{R}^{L_g}$. Here, $L_t$, $L_s$, and $L_g$ denote the feature dimensions of $F_{temp}$, $F_{spat}$, and $F_{goal}$, respectively.

The subsequent multi-head cross-attention (MHCA) module then treats $F_{goal}$ as a query over the concatenated memory $[F_{temp}, F_{spat}]$, yielding the goal-conditioned environmental representation $z_t = \operatorname{MHCA}(F_{goal}, [F_{temp}, F_{spat}])$. The resulting latent representation is then supplied to the downstream policy and value networks.

\paragraph{Policy and Value Networks}
The policy and value networks are implemented as MLPs with ReLU activations. The policy network $\pi$ receives the latent representation $z_t$ concatenated with the ego motion $m_t$ and outputs a predicted action $\hat{a}_t$ with a final tanh squashing. For brevity, we denote it by $\hat{a} = \pi(s)$.

Additionally, we employ twin value networks $Q_1$ and $Q_2$ that process the latent representation $z_t$, the ego motion $m_t$, and the action $a_t$. Their outputs are abbreviated as $\hat{q}_1 = Q_1(s, a)$ and $\hat{q}_2 = Q_2(s, a)$, with value estimation derived from the minimum of these two predictions.

\subsection{Offline RL Training}
The proposed offline RL framework comprises two training stages. In the first stage, we perform Behavior Cloning (BC) to pretrain the state encoder. Subsequently, we employ Conservative Q-Learning (CQL) \cite{kumar2020conservative} to optimize the downstream policy and value networks.

\label{sec:policy optimization}
\paragraph{State Encoder Pretraining} Inspired by Behavior Prior Representation (BPR) \cite{zang2022behavior}, we pretrain the state encoder to facilitate efficient policy optimization. Specifically, an action head $f$ is appended to the encoder, and the network is trained to minimize the BC loss:
\begin{equation}
\min\;
\mathbb{E}_{(s,a)\sim\mathcal{D}}
    \left[\left\lVert
        f\bigl(\phi(s)\bigr) - a
    \right\rVert_2\right].
\end{equation}

By imitating the decisions of the behavior policy, the encoder is encouraged to learn behavior-salient representations. Notably, unlike the original BPR which keeps the encoder frozen, we allow it to be fine-tuned during later training.

\paragraph{Policy Optimization}
During policy optimization, we iteratively update the value networks and the policy network. At each iteration, each value network is updated to minimize the standard temporal-difference (TD) loss augmented with a conservative regularization term that penalizes overestimation of Q-values for out-of-distribution actions:
\begin{equation}
\resizebox{\linewidth}{!}{$
\begin{aligned}
\min\;
\mathbb{E}_{(s,a,r,s')\sim\mathcal{D}}
    \Bigl[
        \bigl(Q(s,a) - \bigl(r+\gamma\min_{j=1,2} Q_{j}\bigl(s',\Pi(s')\bigr)\bigr)\bigr)^{2}
    \Bigr] \\
+\alpha\Bigl(
      \mathbb{E}_{s\sim\mathcal{D}}
        \Bigl[
            \tau\log\sum_{a'\sim{\mu}}\exp\bigl(Q(s,a')/\tau\bigr)
        \Bigr]
    -\mathbb{E}_{(s,a)\sim\mathcal{D}}
        \Bigl[Q(s,a)\Bigr]
\Bigr),
\end{aligned}
$}
\end{equation}
where the first term corresponds to the TD loss, in which $\Pi(s')$ denotes the discrete projection of the continuous target action $\pi(s')$ onto the action space $\mathcal{A}$ as used in Wolpertinger training \cite{dulac2015deep}. The second term is a conservative regularizer, where $\alpha > 0$ weights the strength of conservatism and $\tau > 0$ controls the softness of the log-sum-exp operator. The proposal distribution $\mu$ draws candidate actions by mixing random actions with those generated by the current policy. This term helps suppress over-optimistic Q-values for candidate actions while preserving accurate estimates for actions that are well-supported by the dataset. 

Subsequently, the policy network is updated to maximize the expected Q-values of its actions:

\begin{equation}
\max\;
\mathbb{E}_{s\sim\mathcal{D}}
    \Bigl[
        \min_{j=1,2} 
            Q_{j}\bigl(s,\pi(s)\bigr)
    \Bigr].
\end{equation}

\paragraph{Reward Function}
To guide the RL agent towards safe and efficient parking behaviors, we formulate the total reward as a sum of four terms, including the goal reward $r_g$, the progress reward $r_p$, the collision penalty $r_c$, and the unsafe behavior penalty $r_u$:
\begin{equation}
r = r_g + r_p + r_c + r_u.
\end{equation}

\textbf{Goal Reward $r_g$:} A terminal bonus is granted only when the vehicle stops within the prescribed pose tolerances of the target parking slot:
\begin{equation}
r_g=
\begin{cases}
c_g, & \text{if } p_t\le\varepsilon_p, \\
0, & \text{otherwise},
\end{cases}
\end{equation}
where $c_g > 0$ denotes the reward value for successful completion and $\varepsilon_p$ is the tolerance threshold.

\textbf{Progress Reward $r_p$:} The progress reward encourages the vehicle to move closer to the target at each step. We define it as an exponentially decaying function of the normalized distance to the target:
\begin{equation}
r_p=\exp(-\lambda d_t/d_{total}),
\end{equation}
where $\lambda > 0$ controls the decay rate, $d_t$ is the SE(2) distance from the current pose to the target pose, and $d_{total}$ is the initial distance measured at episode start.

\textbf{Collision Penalty $r_c$:} A constant penalty is applied if the vehicle collides with any obstacle during parking:
\begin{equation}
r_c=
\begin{cases}
-c_c, & \text{if a collision occurs}, \\
0, & \text{otherwise},
\end{cases}
\end{equation}
where $c_c > 0$ determines the penalty strength for collision.

\textbf{Unsafe Behavior Penalty $r_u$:} Similar to prior work \cite{kochdumper2023provably, dunlap2023run}, we introduce a constant penalty $c_u > 0$ to shape the reward whenever a safety constraint is violated. Specifically, we impose a penalty if the current action $a_t$ deviates significantly from the expert action $a^*$, defined as:
\begin{equation}
r_u=
\begin{cases}
-c_u, & \text{if } \lVert a_t - a^*\rVert_2 \ge \varepsilon_a, \\
0, & \text{otherwise},
\end{cases}
\end{equation}
where $\varepsilon_a$ is the predefined threshold for action deviation.

\section{EXPERIMENTS}
\subsection{Experimental Settings}
\paragraph{Implementation Details} All experiments are run on a workstation running Ubuntu 22.04, equipped with a single NVIDIA A100 GPU with 40 GB of memory and 128 GB of RAM. The discrete action space is configured with $N_1$ = 11, $N_2$ = 11, and $N_3$ = 2. Training proceeds in two stages, with 20 epochs of state encoder pretraining and 80 epochs of RL policy optimization. Mini-batches of 256 transitions are used to update the networks with the Adam optimizer. During state encoder pretraining, the encoder uses a learning rate of $1 \times 10^{-4}$ with weight decay $1 \times 10^{-5}$. During RL policy optimization, the actor and each critic are trained with learning rates of $1 \times 10^{-4}$ and $3 \times 10^{-4}$, respectively, while the state encoder is fine-tuned with a learning rate of $1 \times 10^{-5}$. The remaining hyperparameters include a discount factor $\gamma = 0.95$ and CQL coefficients $\alpha = \tau = 1.0$.

\paragraph{Comparison Methods} We benchmark the proposed method against several baselines. The off-policy RL algorithm Soft Actor–Critic (SAC) \cite{haarnoja2018soft}, which we adapt to the purely offline setting, is included to evaluate the benefits of offline RL. In addition, we implement three representative model-free offline RL algorithms from distinct methodological families \cite{prudencio2023survey}, including SAC-$N$ \cite{an2021uncertainty}, BC, and TD3-BC \cite{fujimoto2021minimalist}. SAC-$N$ is an uncertainty estimation method that penalizes Q-values using the variance from an ensemble of value networks. BC is an imitation learning method that learns a policy by directly mimicking the state-action mappings present in the dataset. TD3-BC is a policy constraint method that regularizes the learned policy toward the dataset behavior via BC regularization.

\paragraph{Evaluation Metrics} Performance is quantified using the following metrics.

\textbf{Target Success Rate (TSR):} The fraction of episodes in which the agent reaches the target pose with position error $<$1.2$\,\mathrm{m}$ and orientation error $<$15$\,^\circ$ while avoiding both collisions and timeouts.

\textbf{Target Failure Rate (TFR):} The fraction of episodes in which the agent fails to meet the above target pose tolerances without colliding or timing out.

\textbf{Collision Rate (CR):} The fraction of episodes in which the agent collides with any obstacle.

\textbf{Timeout Rate (TR):} The fraction of episodes in which the agent fails to reach the target pose within the maximum allowed time.

Additionally, for episodes without colliding or timing out, we report three supplementary metrics:

\textbf{Average Position Error (APE):} The mean Euclidean distance between the final position and the center of the target slot.

\textbf{Average Orientation Error (AOE):} The mean absolute difference in yaw between the final heading and the target slot orientation.

\textbf{Average Parking Time (APT):} The mean time taken to complete the parking maneuver.

\begin{figure}[t]
  \centering  
  \includegraphics[width=0.9\columnwidth]{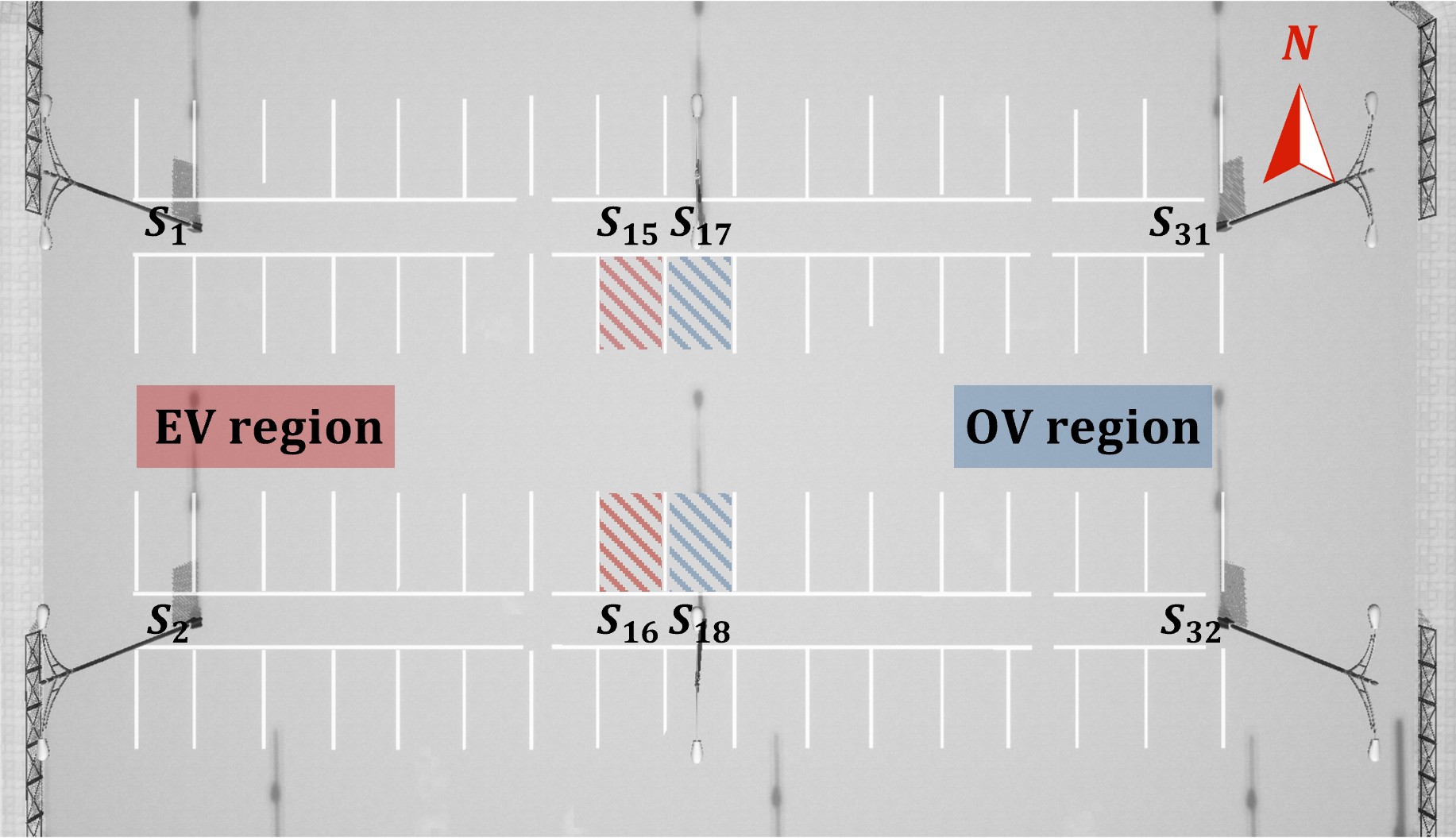}
  \caption{Top view of the parking lot. The initial regions of the EV and OV are represented by solid red and blue rectangles, while their target slots during data collection are indicated by red and blue boxes with diagonal lines respectively.}
  \label{fig:parking-lot}
\end{figure}

\begin{figure}[t]
  \centering
  \begin{subfigure}[b]{0.32\columnwidth}
    \centering
    \includegraphics[width=\linewidth]{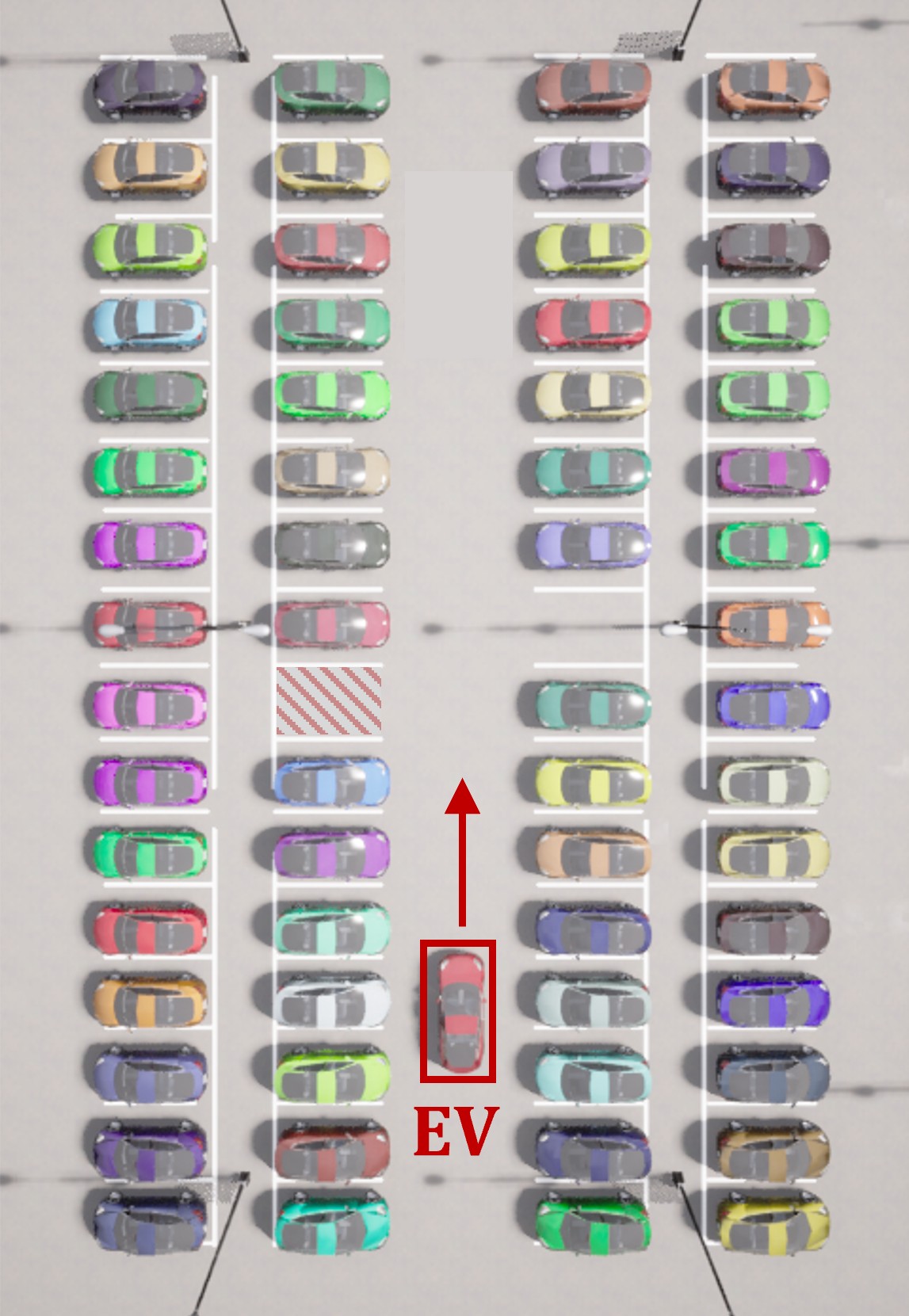}
    \caption{Type \emph{i}}
  \end{subfigure}\hfill
  \begin{subfigure}[b]{0.32\columnwidth}
    \centering
    \includegraphics[width=\linewidth]{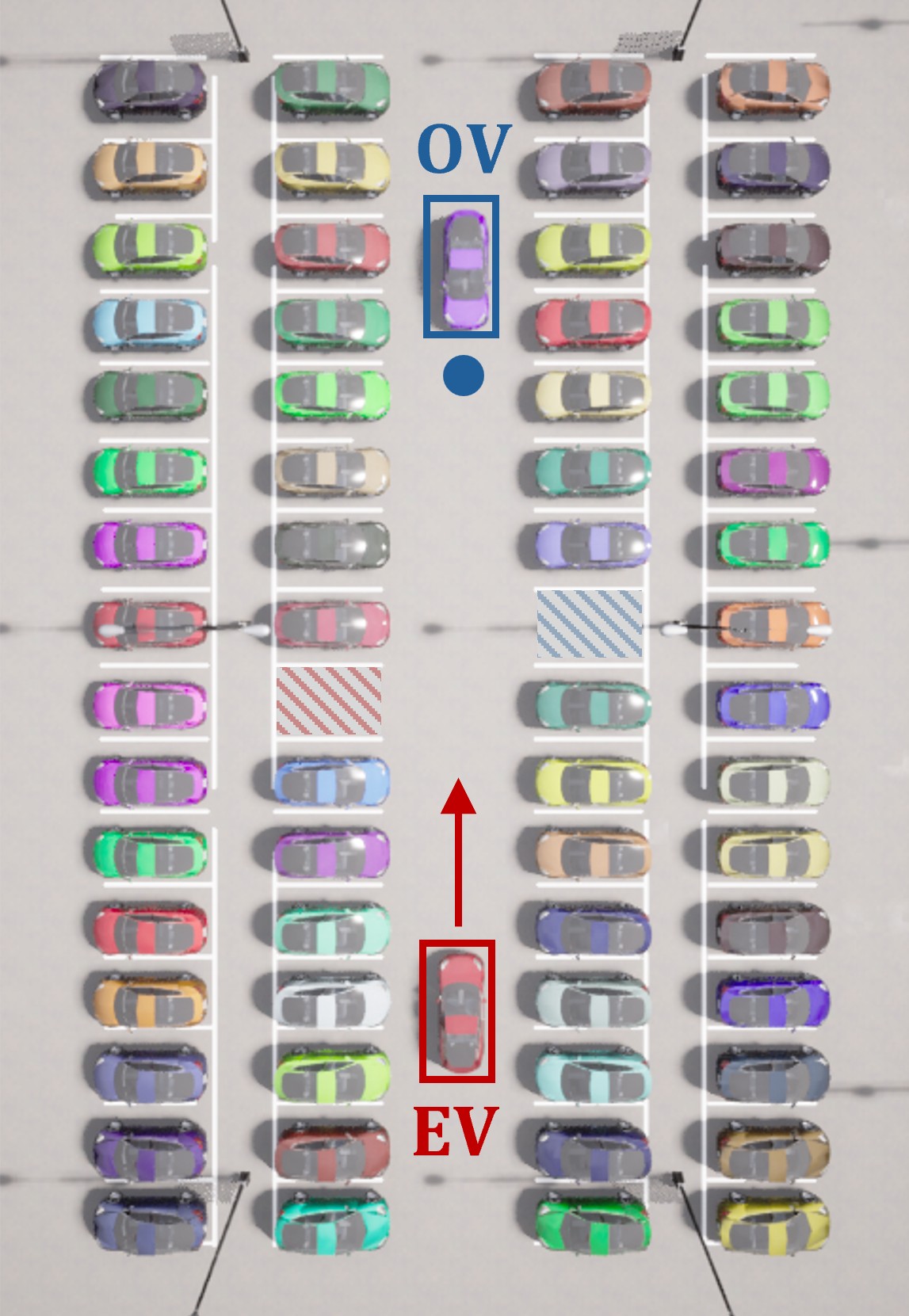}
    \caption{Type \emph{ii}}
  \end{subfigure}\hfill
  \begin{subfigure}[b]{0.32\columnwidth}
    \centering
    \includegraphics[width=\linewidth]{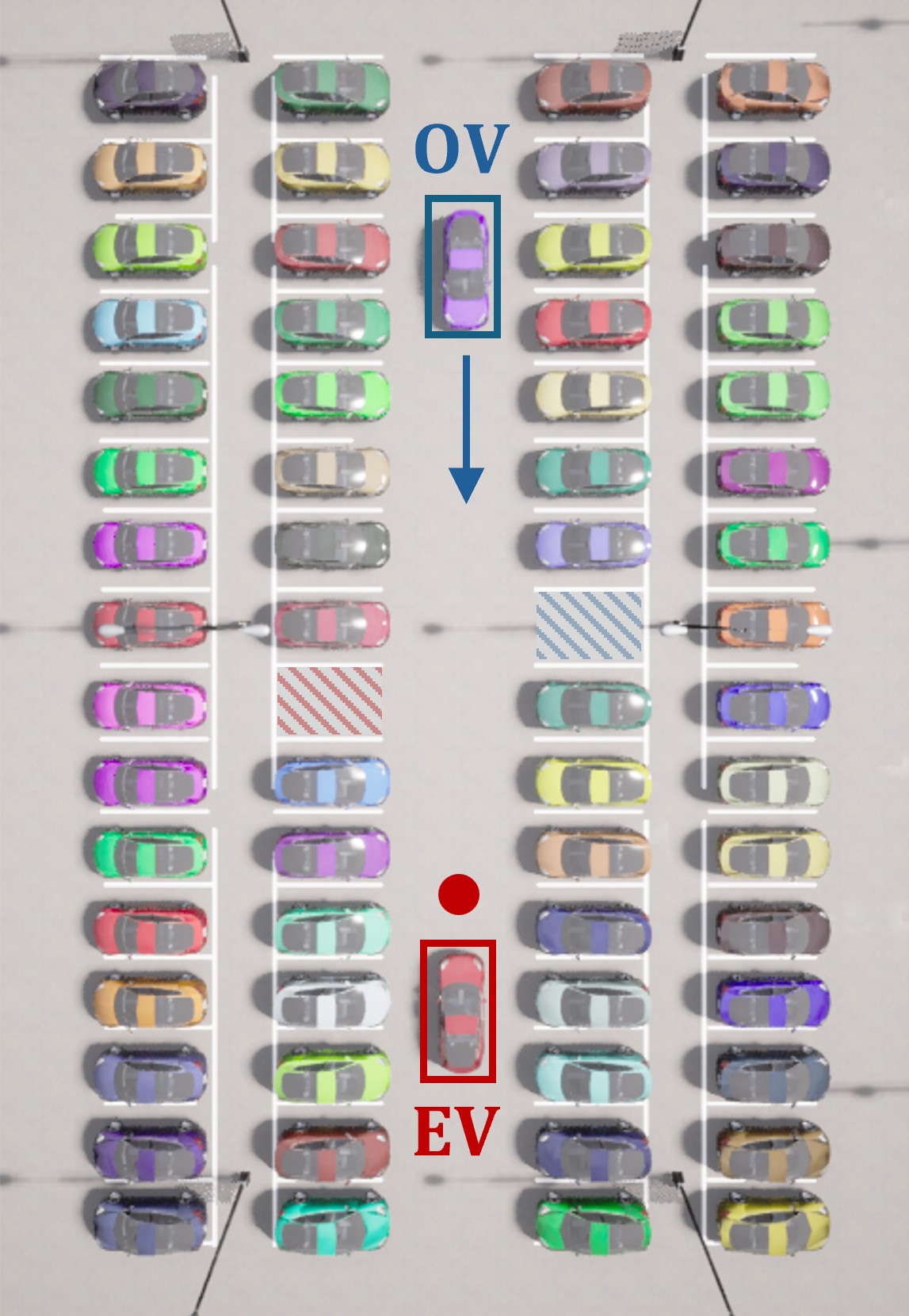}
    \caption{Type \emph{iii}}
  \end{subfigure}
  \caption{Schematic diagram of different parking task types. The EV and OV are highlighted with red and blue solid boxes, respectively. Arrows and dots indicate priority passage and temporary waiting states, respectively.}
  \label{fig:scenario-type}
\end{figure}

\begin{table*}[t]
\centering
\caption{Performance of different methods across three task types on target slots $\{S_{15}, S_{16}\}$}
\label{tab:in-distribution}
\resizebox{0.95\textwidth}{!}{
\begin{tabular}{ccccccccc}
\toprule
Method & Task Type
& TSR (\%) $\uparrow$ & TFR (\%) $\downarrow$ & CR (\%) $\downarrow$ & TR (\%) $\downarrow$ 
& APE (m) $\downarrow$ & AOE (deg) $\downarrow$ & APT (s) $\downarrow$ \\
\midrule
\multirow{3}{*}{SAC}     & \emph{i}   & 2.08  & 64.58 & 33.33 & 0.00  & 1.777 & 30.776 & 21.666 \\
                         & \emph{ii}  & 0.00  & 64.58 & 33.33 & 2.08  & 1.572 & 27.905 & 22.409 \\
                         & \emph{iii} & 0.00  & 6.25  & 93.75 & 0.00  & 0.902 & 16.537 & 28.978 \\
\cmidrule(lr){1-9}
\multirow{3}{*}{SAC-$N$ ($N=20$)} & \emph{i}   & 37.50  & 43.75 & 18.75  & 0.00  & 1.002 & 11.156 & 23.038 \\
                                  & \emph{ii}  & 45.83  & 50.00 & 4.17   & 0.00  & 0.919 & 11.810 & 23.244 \\
                                  & \emph{iii} & 45.83  & 8.33  & 45.83  & 0.00  & 0.816 & 8.109  & 37.373 \\
\cmidrule(lr){1-9}
\multirow{3}{*}{BC}      & \emph{i}   & 60.42 & 0.00  & 20.83 & 18.75 & 0.597 & 7.296 & 38.463 \\
                         & \emph{ii}  & 62.50 & 4.17  & 20.83 & 12.50 & 0.628 & 6.970 & 38.357 \\
                         & \emph{iii} & 66.67 & 0.00  & 22.92 & 10.42 & 0.653 & 6.160 & 51.794 \\
\cmidrule(lr){1-9}
\multirow{3}{*}{TD3-BC}  & \emph{i}   & 75.00  & 22.92 & 2.08  & 0.00  & 0.624 & 9.674 & 27.423 \\
                         & \emph{ii}  & 85.42  & 14.58 & 0.00  & 0.00  & 0.535 & 8.153 & 27.802 \\
                         & \emph{iii} & 72.92  & 4.17  & 22.92 & 0.00  & 0.396 & 6.657 & 41.450 \\
\cmidrule(lr){1-9}
\multirow{3}{*}{Ours}    & \emph{i}   & 97.92  & 2.08  & 0.00  & 0.00  & 0.576 & 8.675  & 26.165 \\
                         & \emph{ii}  & 91.67  & 4.17  & 4.17  & 0.00  & 0.475 & 7.925  & 26.181 \\
                         & \emph{iii} & 97.92  & 2.08  & 0.00  & 0.00  & 0.526 & 10.511 & 40.962 \\
\bottomrule
\end{tabular}
}
\end{table*}

\begin{table*}[t]
\centering
\caption{Performance of different methods across three task types on target slots $\{S_{11}, S_{12}\}$}
\label{tab:out-of-distribution}
\resizebox{0.95\textwidth}{!}{
\begin{tabular}{ccccccccc}
\toprule
Method & Task Type 
& TSR (\%) $\uparrow$ & TFR (\%) $\downarrow$ & CR (\%) $\downarrow$ & TR (\%) $\downarrow$ 
& APE (m) $\downarrow$ & AOE (deg) $\downarrow$ & APT (s) $\downarrow$ \\
\midrule
\multirow{3}{*}{SAC}     & \emph{i}   & 2.08  & 10.42 & 87.50 & 0.00  & 2.159 & 36.915 & 15.322 \\
                         & \emph{ii}  & 6.25  & 10.42 & 83.33 & 0.00  & 1.859 & 34.184 & 20.475 \\
                         & \emph{iii} & 2.08  & 2.08  & 95.83 & 0.00  & 1.830 & 30.944 & 17.067 \\
\cmidrule(lr){1-9}
\multirow{3}{*}{SAC-$N$ ($N=20$)} & \emph{i}   & 47.92  & 6.25  & 45.83  & 0.00  & 0.588 & 6.244 & 17.383 \\
                                  & \emph{ii}  & 58.33  & 22.92 & 18.75  & 0.00  & 0.616 & 9.338 & 18.278 \\
                                  & \emph{iii} & 16.67  & 2.08  & 81.25  & 0.00  & 0.567 & 5.855 & 29.941 \\
\cmidrule(lr){1-9}
\multirow{3}{*}{BC}      & \emph{i}   & 2.08  & 0.00  & 97.92 & 0.00  & 0.528 & 9.743 & 29.733 \\
                         & \emph{ii}  & 4.17  & 0.00  & 93.75 & 2.08  & 0.885 & 7.555 & 33.967 \\
                         & \emph{iii} & 6.25  & 0.00  & 93.75 & 0.00  & 0.784 & 5.883 & 33.422 \\
\cmidrule(lr){1-9}
\multirow{3}{*}{TD3-BC}  & \emph{i}   & 18.75  & 0.00  & 81.25  & 0.00  & 0.434 & 5.787 & 22.141 \\
                         & \emph{ii}  & 39.58  & 2.08  & 58.33  & 0.00  & 0.442 & 4.523 & 22.925 \\
                         & \emph{iii} & 20.83  & 0.00  & 79.17  & 0.00  & 0.336 & 3.810 & 33.033 \\
\cmidrule(lr){1-9}
\multirow{3}{*}{Ours}    & \emph{i}   & 91.67  & 0.00  & 8.33  & 0.00  & 0.302 & 6.483 & 22.104 \\
                         & \emph{ii}  & 91.67  & 0.00  & 4.17  & 4.17  & 0.269 & 4.245 & 24.223 \\
                         & \emph{iii} & 97.92  & 0.00  & 2.08  & 0.00  & 0.363 & 7.158 & 36.299 \\
\bottomrule
\end{tabular}
}
\end{table*}

\subsection{Dataset Collection}
The parking maneuver dataset is collected in CARLA \cite{dosovitskiy2017carla} in the parking lot of the Town04\_Opt map, as shown in Fig.~\ref{fig:parking-lot}. The parking slots in the two central rows, each containing 16 slots, are indexed as $\{S_i\mid i=1, 2, ...,32\}$. 

In each episode, the EV is initialized uniformly within a 12.0$\,\mathrm{m}$ $\times$ 2.5$\,\mathrm{m}$ rectangular region at the western end of the central aisle, oriented eastward. Its target slot is chosen from $\{S_{15}, S_{16}\}$. When the OV is present, it is initialized uniformly within an identically sized region at the eastern end of the same aisle, oriented westward, and its target slot is drawn independently from $\{S_{17}, S_{18}\}$. All non-target slots are occupied by static vehicles that serve as obstacles.

Based on the presence of the OV and its right-of-way relative to the EV, we consider three parking task types. First, no OV is present, and only the EV performs the parking maneuver. Second, an OV is present with low priority and yields to the EV. Third, an OV is present with high priority, so it proceeds first and the EV must wait. These three types are visualized in Fig.~\ref{fig:scenario-type}. For both the EV and OV, the reference path is planned using the Hybrid A* algorithm and tracked by an LQR controller. At each control step of the EV, the expert action is executed with probability 0.8. Otherwise, with probability 0.2, a random control command is applied. Each episode terminates when the EV reaches the target slot within a specified pose tolerance or a collision occurs. A total of 100 episodes are collected for each task type.

\subsection{Evaluation Results}
The closed-loop experiments are conducted in CARLA to evaluate agents trained by different methods across three task types. In each evaluation, the agent is required to complete the parking maneuver within time limits of 45$\,\mathrm{s}$ for types \emph{i} and \emph{ii}, and 60$\,\mathrm{s}$ for type \emph{iii}. For each task type and target slot, 24 evaluation episodes are conducted with different starting positions to ensure statistical robustness.

The in-distribution performance is first evaluated using the target slots $\{S_{15}, S_{16}\}$. Table~\ref{tab:in-distribution} reports the performance of different methods across three task types on these in-distribution slots. Our method achieves the highest TSR across all task types, along with the lowest CR and no recorded timeouts. By contrast, a naive implementation of SAC without conservative regularization fails to learn an effective policy due to extrapolation error, while the ensemble-based variant SAC-$N$ alleviates this issue by estimating uncertain Q-values. Notably, we observe that both SAC and SAC-$N$ tend to select more aggressive actions during parking, resulting in a shorter APT but higher CR and larger pose errors. On the other hand, since the pre-collected dataset is not entirely composed of expert demonstrations, BC achieves a relatively low TSR and a notable proportion of timeouts as it blindly mimics all behaviors without knowledge of the rewards. Although TD3-BC narrows this gap by integrating policy evaluation and improvement, it still fails to strike an effective balance between policy optimization and behavioral imitation.

\begin{figure*}[t]
  \centering
  \includegraphics[width=\textwidth]{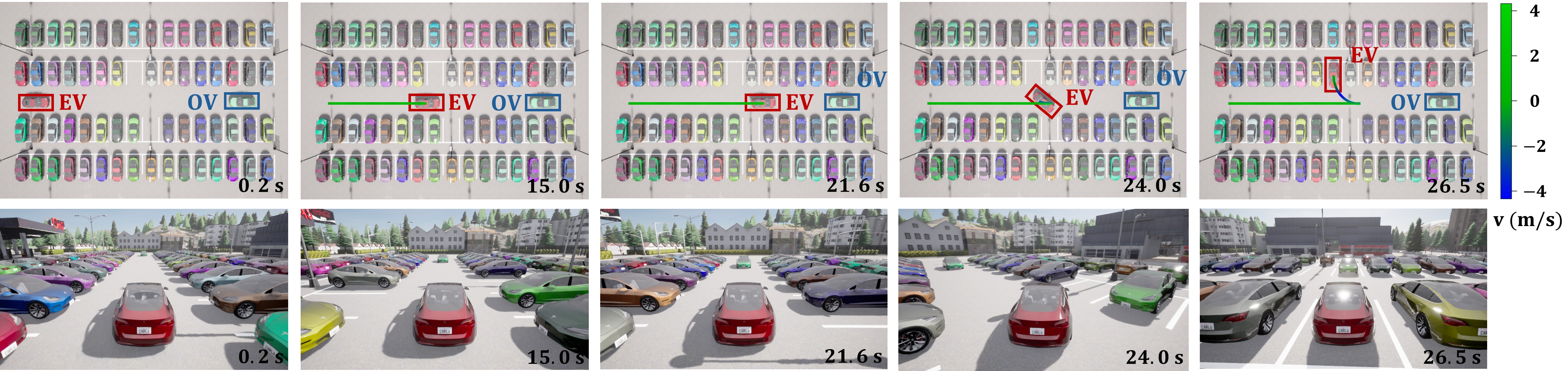}
  \caption{Key frames of the interactive parking process under task type \emph{ii} on target slot $S_{15}$. The top section presents a top view, in which the parking trajectory of the EV is represented by a solid line whose color varies according to speed. The bottom section provides a third-person perspective view of the EV.}
  \label{fig:demo_ID}
\end{figure*}

\begin{figure*}[t]
  \centering
  \includegraphics[width=\textwidth]{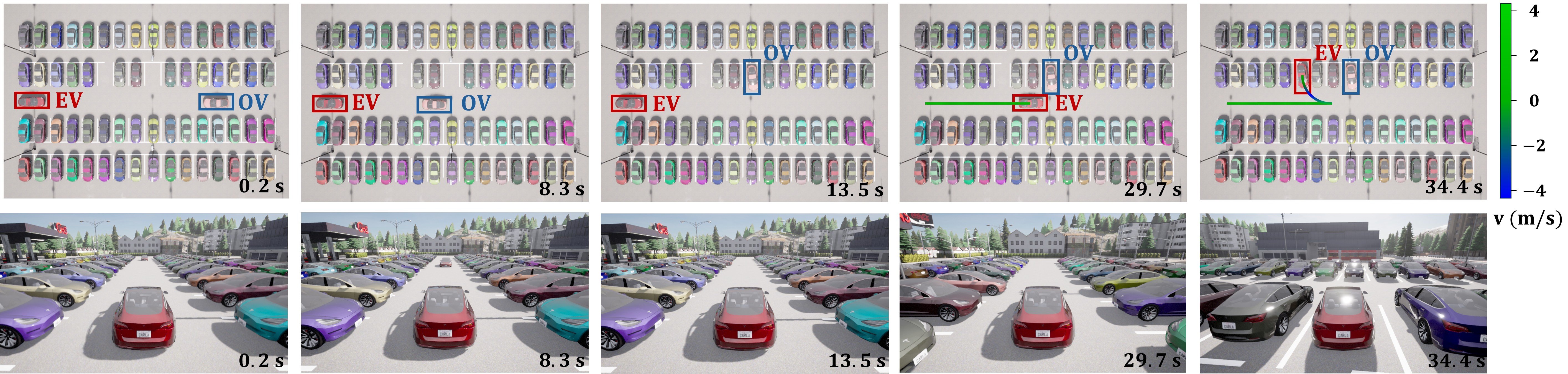}
  \caption{Key frames of the interactive parking process under task type \emph{iii} on target slot $S_{11}$.}
  \label{fig:demo_OOD}
\end{figure*}

In addition to the in-distribution slots, two out-of-distribution target slots $\{S_{11}, S_{12}\}$ are used to validate the generalization ability of each method. The results are summarized in Table~\ref{tab:out-of-distribution}. Our method demonstrates strong adaptability to unseen parking scenarios, maintaining a high TSR and low CR. This indicates that the proposed method can make appropriate decisions when the surrounding environment changes. In comparison, both BC and TD3-BC suffer significant performance degradation with a notably increased CR due to the distribution shift. SAC-$N$, though less severely impacted by the shift, still exhibits subpar performance when evaluated on the unseen slots. The comparative results demonstrate that the performance of our method can be attributed to its penalization for potentially unsafe candidate actions, leading it to collapse to good-enough decisions in parking scenarios.

\subsection{Qualitative Demonstrations}
To better illustrate the performance of our method, we select one representative evaluation episode from task type \emph{ii} and another from task type \emph{iii} for visualization. The key frames of these two interactive parking processes are shown in Fig.~\ref{fig:demo_ID} and Fig.~\ref{fig:demo_OOD}, respectively. For the type \emph{ii} episode, the target slot is assigned to $S_{15}$. At the beginning, the EV perceives that the OV is waiting in place and therefore decides to initiate the parking maneuver. The EV proceeds straight until it reaches the turning point at 21.6$\,\mathrm{s}$. It then shifts into reverse gear and, within the next 4.9$\,\mathrm{s}$, performs a smooth backing maneuver with coordinated steering. Finally, the EV successfully completes the parking task.

For the type \emph{iii} episode, the target slot is assigned to an out-of-distribution slot $S_{11}$. In this episode, the OV proceeds first, so the EV decides to remain stationary and yield. At 13.5$\,\mathrm{s}$, once the EV perceives that the OV has completed its parking maneuver, it begins to move and reaches the turning point at 29.7$\,\mathrm{s}$. After 4.7$\,\mathrm{s}$ of precise backing operation, it also ideally enters the designated target slot. These two visualizations highlight the effectiveness and generalization capability of the proposed method.

\section{CONCLUSIONS}
In this work, we propose an innovative end-to-end offline RL framework for autonomous parking that learns safe and efficient parking policies directly from a pre-collected dataset. Our method integrates a goal-conditioned state encoder pretrained to distill behavior priors, and a downstream policy optimized with conservative regularization to mitigate extrapolation error. To support training and evaluation, we construct a novel parking dataset that encompasses interactive scenarios with the OV. Extensive experiments conducted in the high-fidelity CARLA simulator demonstrate the effectiveness of the proposed method. Our method achieves a high parking success rate with few collisions across different task types. Moreover, it exhibits strong generalization capability to unseen target slots, outperforming several baselines. While this work is limited to simulation at the current stage, the results validate the potential of offline RL as a promising paradigm for safe and scalable autonomous parking systems.

\bibliographystyle{IEEEtran}
\bibliography{refs}

\begin{thebibliography}{10}
\providecommand{\url}[1]{#1}
\csname url@samestyle\endcsname
\providecommand{\newblock}{\relax}
\providecommand{\bibinfo}[2]{#2}
\providecommand{\BIBentrySTDinterwordspacing}{\spaceskip=0pt\relax}
\providecommand{\BIBentryALTinterwordstretchfactor}{4}
\providecommand{\BIBentryALTinterwordspacing}{\spaceskip=\fontdimen2\font plus
\BIBentryALTinterwordstretchfactor\fontdimen3\font minus \fontdimen4\font\relax}
\providecommand{\BIBforeignlanguage}[2]{{%
\expandafter\ifx\csname l@#1\endcsname\relax
\typeout{** WARNING: IEEEtran.bst: No hyphenation pattern has been}%
\typeout{** loaded for the language `#1'. Using the pattern for}%
\typeout{** the default language instead.}%
\else
\language=\csname l@#1\endcsname
\fi
#2}}
\providecommand{\BIBdecl}{\relax}
\BIBdecl

\bibitem{zhao2024survey}
R.~Zhao, Y.~Li, Y.~Fan, F.~Gao, M.~Tsukada, and Z.~Gao, ``A survey on recent advancements in autonomous driving using deep reinforcement learning: Applications, challenges, and solutions,'' \emph{IEEE Transactions on Intelligent Transportation Systems}, 2024.

\bibitem{li2021optimization}
B.~Li, T.~Acarman, Y.~Zhang, Y.~Ouyang, C.~Yaman, Q.~Kong, X.~Zhong, and X.~Peng, ``Optimization-based trajectory planning for autonomous parking with irregularly placed obstacles: A lightweight iterative framework,'' \emph{IEEE Transactions on Intelligent Transportation Systems}, vol.~23, no.~8, pp. 11\,970--11\,981, 2021.

\bibitem{leu2022autonomous}
J.~Leu, Y.~Wang, M.~Tomizuka, and S.~Di~Cairano, ``Autonomous vehicle parking in dynamic environments: An integrated system with prediction and motion planning,'' in \emph{International Conference on Robotics and Automation}, 2022, pp. 10\,890--10\,897.

\bibitem{sheng2021autonomous}
W.~Sheng, B.~Li, and X.~Zhong, ``Autonomous parking trajectory planning with tiny passages: A combination of multistage hybrid a-star algorithm and numerical optimal control,'' \emph{IEEE Access}, vol.~9, pp. 102\,801--102\,810, 2021.

\bibitem{guo2025fast}
Z.~Guo, Y.~Wang, H.~Yu, and J.~Xi, ``Fast optimization-based trajectory planning with cumulative key constraints for automated parking in unstructured environments,'' \emph{IEEE Transactions on Vehicular Technology}, 2025.

\bibitem{dai2018improving}
S.~Dai, M.~Orton, S.~Schaffert, A.~Hofmann, and B.~Williams, ``Improving trajectory optimization using a roadmap framework,'' in \emph{IEEE/RSJ International Conference on Intelligent Robots and Systems}, 2018, pp. 8674--8681.

\bibitem{codevilla2018end}
F.~Codevilla, M.~M{\"u}ller, A.~L{\'o}pez, V.~Koltun, and A.~Dosovitskiy, ``End-to-end driving via conditional imitation learning,'' in \emph{IEEE International Conference on Robotics and Automation}, 2018, pp. 4693--4700.

\bibitem{de2019causal}
P.~De~Haan, D.~Jayaraman, and S.~Levine, ``Causal confusion in imitation learning,'' \emph{Advances in Neural Information Processing Systems}, vol.~32, 2019.

\bibitem{kumar2022should}
A.~Kumar, J.~Hong, A.~Singh, and S.~Levine, ``When should we prefer offline reinforcement learning over behavioral cloning?'' \emph{arXiv preprint arXiv:2204.05618}, 2022.

\bibitem{kiran2021deep}
B.~R. Kiran, I.~Sobh, V.~Talpaert, P.~Mannion, A.~A. Al~Sallab, S.~Yogamani, and P.~P{\'e}rez, ``Deep reinforcement learning for autonomous driving: A survey,'' \emph{IEEE Transactions on Intelligent Transportation Systems}, vol.~23, no.~6, pp. 4909--4926, 2021.

\bibitem{peng2025bilevel}
Z.~Peng, Y.~Wang, L.~Zheng, and J.~Ma, ``Bilevel multi-armed bandit-based hierarchical reinforcement learning for interaction-aware self-driving at unsignalized intersections,'' \emph{IEEE Transactions on Vehicular Technology}, vol.~74, no.~6, pp. 8824--8838, 2025.

\bibitem{peng2024reward}
Z.~Peng, X.~Zhou, L.~Zheng, Y.~Wang, and J.~Ma, ``Reward-driven automated curriculum learning for interaction-aware self-driving at unsignalized intersections,'' in \emph{IEEE/RSJ International Conference on Intelligent Robots and Systems}, 2024.

\bibitem{jiang2025hope}
M.~Jiang, Y.~Li, S.~Zhang, S.~Chen, C.~Wang, and M.~Yang, ``{HOPE}: A reinforcement learning-based hybrid policy path planner for diverse parking scenarios,'' \emph{IEEE Transactions on Intelligent Transportation Systems}, 2025.

\bibitem{gulcehre2020rl}
C.~Gulcehre, Z.~Wang, A.~Novikov, T.~Paine, S.~G{\'o}mez, K.~Zolna, R.~Agarwal, J.~S. Merel, D.~J. Mankowitz, C.~Paduraru \emph{et~al.}, ``{RL} unplugged: A suite of benchmarks for offline reinforcement learning,'' \emph{Advances in Neural Information Processing Systems}, vol.~33, pp. 7248--7259, 2020.

\bibitem{tang2025deep}
C.~Tang, B.~Abbatematteo, J.~Hu, R.~Chandra, R.~Mart{\'\i}n-Mart{\'\i}n, and P.~Stone, ``Deep reinforcement learning for robotics: A survey of real-world successes,'' in \emph{Proceedings of the AAAI Conference on Artificial Intelligence}, vol.~39, no.~27, 2025, pp. 28\,694--28\,698.

\bibitem{dubins1957curves}
L.~E. Dubins, ``On curves of minimal length with a constraint on average curvature, and with prescribed initial and terminal positions and tangents,'' \emph{American Journal of Mathematics}, vol.~79, no.~3, pp. 497--516, 1957.

\bibitem{reeds1990optimal}
J.~Reeds and L.~Shepp, ``Optimal paths for a car that goes both forwards and backwards,'' \emph{Pacific Journal of Mathematics}, vol. 145, no.~2, pp. 367--393, 1990.

\bibitem{kim2014auto}
J.~M. Kim, K.~I. Lim, and J.~H. Kim, ``Auto parking path planning system using modified reeds-shepp curve algorithm,'' in \emph{International Conference on Ubiquitous Robots and Ambient Intelligence}, 2014, pp. 311--315.

\bibitem{du2014autonomous}
X.~Du and K.~K. Tan, ``Autonomous reverse parking system based on robust path generation and improved sliding mode control,'' \emph{IEEE Transactions on Intelligent Transportation Systems}, vol.~16, no.~3, pp. 1225--1237, 2014.

\bibitem{dolgov2010path}
D.~Dolgov, S.~Thrun, M.~Montemerlo, and J.~Diebel, ``Path planning for autonomous vehicles in unknown semi-structured environments,'' \emph{International Journal of Robotics Research}, vol.~29, no.~5, pp. 485--501, 2010.

\bibitem{sedighi2019guided}
S.~Sedighi, D.-V. Nguyen, and K.-D. Kuhnert, ``Guided hybrid a-star path planning algorithm for valet parking applications,'' in \emph{International Conference on Control, Automation and Robotics}, 2019, pp. 570--575.

\bibitem{zips2016optimisation}
P.~Zips, M.~B{\"o}ck, and A.~Kugi, ``Optimisation based path planning for car parking in narrow environments,'' \emph{Robotics and Autonomous Systems}, vol.~79, pp. 1--11, 2016.

\bibitem{chen2024end}
L.~Chen, P.~Wu, K.~Chitta, B.~Jaeger, A.~Geiger, and H.~Li, ``End-to-end autonomous driving: Challenges and frontiers,'' \emph{IEEE Transactions on Pattern Analysis and Machine Intelligence}, 2024.

\bibitem{rathour2018vision}
S.~Rathour, V.~John, M.~Nithilan, and S.~Mita, ``Vision and dead reckoning-based end-to-end parking for autonomous vehicles,'' in \emph{IEEE Intelligent Vehicles Symposium}, 2018, pp. 2182--2187.

\bibitem{li2018end}
R.~Li, W.~Wang, Y.~Chen, S.~Srinivasan, and V.~N. Krovi, ``An end-to-end fully automatic bay parking approach for autonomous vehicles,'' in \emph{Dynamic Systems and Control Conference}, vol. 51906, 2018, p. V002T15A004.

\bibitem{yang2024e2e}
Y.~Yang, D.~Chen, T.~Qin, X.~Mu, C.~Xu, and M.~Yang, ``{E2E} parking: Autonomous parking by the end-to-end neural network on the carla simulator,'' in \emph{IEEE Intelligent Vehicles Symposium}, 2024, pp. 2375--2382.

\bibitem{li2024parkinge2e}
C.~Li, Z.~Ji, Z.~Chen, T.~Qin, and M.~Yang, ``Parking{E2E}: Camera-based end-to-end parking network, from images to planning,'' in \emph{IEEE/RSJ International Conference on Intelligent Robots and Systems}, 2024, pp. 13\,206--13\,212.

\bibitem{song2022time}
S.~Song, H.~Chen, H.~Sun, M.~Liu, and T.~Xia, ``Time-optimized online planning for parallel parking with nonlinear optimization and improved monte carlo tree search,'' \emph{IEEE Robotics and Automation Letters}, vol.~7, no.~2, pp. 2226--2233, 2022.

\bibitem{yuan2023hierarchical}
Z.~Yuan, Z.~Wang, X.~Li, L.~Li, and L.~Zhang, ``Hierarchical trajectory planning for narrow-space automated parking with deep reinforcement learning: A federated learning scheme,'' \emph{Sensors}, vol.~23, no.~8, p. 4087, 2023.

\bibitem{kalman1960contributions}
R.~E. Kalman \emph{et~al.}, ``Contributions to the theory of optimal control,'' \emph{Bol. Soc. Mat. Mexicana}, vol.~5, no.~2, pp. 102--119, 1960.

\bibitem{kumar2020conservative}
A.~Kumar, A.~Zhou, G.~Tucker, and S.~Levine, ``Conservative q-learning for offline reinforcement learning,'' \emph{Advances in Neural Information Processing Systems}, vol.~33, pp. 1179--1191, 2020.

\bibitem{fujimoto2019benchmarking}
S.~Fujimoto, E.~Conti, M.~Ghavamzadeh, and J.~Pineau, ``Benchmarking batch deep reinforcement learning algorithms,'' \emph{arXiv preprint arXiv:1910.01708}, 2019.

\bibitem{gulcehre2021regularized}
C.~Gulcehre, S.~G. Colmenarejo, Z.~Wang, J.~Sygnowski, T.~Paine, K.~Zolna, Y.~Chen, M.~Hoffman, R.~Pascanu, and N.~de~Freitas, ``Regularized behavior value estimation,'' \emph{arXiv preprint arXiv:2103.09575}, 2021.

\bibitem{zang2022behavior}
H.~Zang, X.~Li, J.~Yu, C.~Liu, R.~Islam, R.~T.~D. Combes, and R.~Laroche, ``Behavior prior representation learning for offline reinforcement learning,'' \emph{arXiv preprint arXiv:2211.00863}, 2022.

\bibitem{dulac2015deep}
G.~Dulac-Arnold, R.~Evans, H.~van Hasselt, P.~Sunehag, T.~Lillicrap, J.~Hunt, T.~Mann, T.~Weber, T.~Degris, and B.~Coppin, ``Deep reinforcement learning in large discrete action spaces,'' \emph{arXiv preprint arXiv:1512.07679}, 2015.

\bibitem{kochdumper2023provably}
N.~Kochdumper, H.~Krasowski, X.~Wang, S.~Bak, and M.~Althoff, ``Provably safe reinforcement learning via action projection using reachability analysis and polynomial zonotopes,'' \emph{IEEE Open Journal of Control Systems}, vol.~2, pp. 79--92, 2023.

\bibitem{dunlap2023run}
K.~Dunlap, M.~Mote, K.~Delsing, and K.~L. Hobbs, ``Run time assured reinforcement learning for safe satellite docking,'' \emph{Journal of Aerospace Information Systems}, vol.~20, no.~1, pp. 25--36, 2023.

\bibitem{haarnoja2018soft}
T.~Haarnoja, A.~Zhou, P.~Abbeel, and S.~Levine, ``Soft actor-critic: Off-policy maximum entropy deep reinforcement learning with a stochastic actor,'' in \emph{International Conference on Machine Learning}, 2018, pp. 1861--1870.

\bibitem{prudencio2023survey}
R.~F. Prudencio, M.~R. Maximo, and E.~L. Colombini, ``A survey on offline reinforcement learning: Taxonomy, review, and open problems,'' \emph{IEEE Transactions on Neural Networks and Learning Systems}, vol.~35, no.~8, pp. 10\,237--10\,257, 2023.

\bibitem{an2021uncertainty}
G.~An, S.~Moon, J.-H. Kim, and H.~O. Song, ``Uncertainty-based offline reinforcement learning with diversified q-ensemble,'' \emph{Advances in Neural Information Processing Systems}, vol.~34, pp. 7436--7447, 2021.

\bibitem{fujimoto2021minimalist}
S.~Fujimoto and S.~S. Gu, ``A minimalist approach to offline reinforcement learning,'' \emph{Advances in Neural Information Processing Systems}, vol.~34, pp. 20\,132--20\,145, 2021.

\bibitem{dosovitskiy2017carla}
A.~Dosovitskiy, G.~Ros, F.~Codevilla, A.~Lopez, and V.~Koltun, ``{CARLA}: An open urban driving simulator,'' in \emph{Conference on Robot Learning}, 2017, pp. 1--16.

\end{thebibliography}

\end{document}